# SS-BERT: Mitigating Identity Terms Bias in Toxic Comment Classification by Utilising the Notion of "Subjectivity" and "Identity Terms"

**Zhixue Zhao**
University of Sheffield
United Kingdom
zhixue.zhao@sheffield.ac.uk

**Ziqi Zhang**
University of Sheffield
United Kingdom
ziqi.zhang@sheffield.ac.uk

**Frank Hopfgartner**
University of Sheffield
United Kingdom
f.hopfgartner@sheffield.ac.uk

September 6, 2021

## Abstract

Toxic comment classification models are often found biased toward identity terms which are terms characterising a specific group of people such as "Muslim" and "black". Such bias is commonly reflected in false positive predictions, i.e. non-toxic comments with identity terms. In this work, we propose a novel approach to tackle such bias in toxic comment classification, leveraging the notion of subjectivity level of a comment and the presence of identity terms. We hypothesize that when a comment is made about a group of people that is characterised by an identity term, the likelihood of that comment being toxic is associated with the subjectivity level of the comment, i.e. the extent to which the comment conveys personal feelings and opinions. Building upon the BERT model, we propose a new structure that is able to leverage these features, and thoroughly evaluate our model on 4 datasets of varying sizes and representing different social media platforms. The results show that our model can consistently outperform BERT and a SOTA model devised to address identity term bias in a different way, with a maximum improvement in F1 of 2.43% and 1.91% respectively.

## 1 Introduction

Combating toxic comments online is an important area of research nowadays. It is commonly handled as a text classification task, known as toxic comment classification (TCC). TCC has taken years of research, moving from the earlier methods based on feature engineering with classic machine learning methods to pre-trained language model-based methods. However, several studies have revealed bias in SOTA approaches on TCC tasks, especially bias toward identity terms, known as identity term bias [1, 2, 3]. Identity terms are words or terms referring to specific groups of people, such as "Muslim", "black", "women" and "democrat". This identity term bias is often associated with false positive bias [2]. The scenario is that when an identity term appears in a non-toxic comment, the model tends to classify it as positive, i.e. a toxic comment.

Limited studies have proposed methods to handle such bias and those methods follow a simple principle: ignoring or paying less attention to the identity terms. However, this overlooks the fact that identity terms can be essential and important features to make predictions. In this work, we explore a new approach: when an identity term appears in a comment, we ask the model to incorporate the level of "subjectivity" in prediction. A comment with a low subjectivity level expresses more factual information and less personal feelings and opinions; while a comment with a high subjectivity level contains more personal opinions but less factual information. We hypothesize that when a toxic comment is made about a group of people and this comment contains identity terms, it is more likely to express subjective opinions than citing factual information. Therefore, The likelihood of a comment containing identity terms to be toxic can be associated with the subjectivity level of that comment.

Building on the BERT model which is commonly used for multiple downstream NLP tasks, we experiment with three different structures that each utilise: subjectivity and identity terms (Subdentity-Sensitive BERT where "Subdentity" denotes "subjectivity" and "identity", thus we call it *SS-BERT* in short), subjectivity alone (Subjectivity-Only BERT,

*SO-BERT*), and subjectivity and identity terms with Sampling and Occlusion (SOC) regularization built on the method by [3] (*SS-BERT+SOC*). We evaluate these models on a wide range of TCC tasks with different dataset sizes, different text lengths and from different social media platforms. Our model achieves consistent improvement over SOTA methods. First, SS-BERT consistently outperforms BERT, particularly in predicting fewer false positives. This indicates that using the subjectivity levels and the presence of identity terms is helpful to mitigate the false positive bias. Second, SS-BERT consistently outperforms SO-BERT and SS-BERT+SOC. This indicates that simply learning the subjectivity level for all comments is not enough, and it is more informative to combine subjectivity with the presence of identity terms. Our main contributions are as follows:

- We analysis the bias towards identity terms found in BERT on toxic comment classification tasks, and identify the relationship between such bias and the subjectivity level of comments on a wide range of datasets.
- We create a BERT-based model which utilizes the relationship mentioned above to handle the identity term bias. Our model achieves consistent improvements across a range of datasets.
- We compare different models for handling identity term bias and further confirm the relationship mentioned above and our model.

## 2 Related work

### 2.1 Toxic comment classification models

Toxic comment generally refers to different types of negative, unhealthy or disrespectful user-generated content online, which includes hate speech, abusive language, cyberbullying, etc [4, 5, 6]. The majority of studies handle TCC as a binary classification problem [7]. A few studies frame TCC as multi-class classification tasks where a comment will be assigned into one of the multiple types of toxic comment, or multi-label tasks where a comment could be assigned into none or one or multiple types of toxic comments [8, 9, 10].

Early research on TCC primarily makes use of traditional machine learning algorithms, such as logistic regression, Support Vector Machines and Naïve Bayes [11]. Since 2010, studies on TCC have shifted towards deep neural networks (DNN)-based models. These include Convolutional Neural Networks, Recurrent Neural Networks, Bi-directional Long Short-Term Memory network and hybrid neural networks which combine different DNN configurations [12, 13, 11, 14].

With the introduction of transformer-based DNN architectures, the use of pre-trained language models (LMs) in downstream TCC tasks has become mainstream [15]. Commonly-used pre-trained LMs include BERT, RoBERTa, XLM, etc. [15, 16, 17]. A few studies of TCC tasks explore these pre-trained LMs. For example, [18] and [19] transfer pre-trained LMs onto different downstream architectures. [19] also test continued pre-training of a LM with in-domain corpus.

### 2.2 Identity term bias

Recent studies show that TCC classifiers often suffer from unintended bias, especially the bias towards "identity terms" (also known as "group identifiers"), which are words or terms referring to people with specific demographic characteristics, such as ethnic origin, religion, gender, or sexual orientation. These studies point out that TCC models tend to assign too much attention to such identity terms, resulting in incorrect predictions [1, 2, 3]. This is called "identity term bias" [2]. Besides, such bias towards identity terms often reflects the false positive predictions, known as false positive bias. For example, [1] give an example in their study that "You are a good woman" is predicted as "sexist".

To handle the bias towards gender identity terms, [1] test de-biasing word embeddings, data augmentation via switching female and male entities, and pre-training the model on a similar and larger but less-biased dataset first. Data augmentation and extensive pre-training both essentially let the model learn from more data.

[2] investigate the data imbalance problem which they argue contributes to the bias towards identity terms. They find that identity terms appear much more often in toxic comments than in non-toxic comments and they believe this imbalance leads the classifier to generalise identity terms as indicative features for toxicity, thus over-predict false positives. To address this issue, they manually add non-toxic comments with identity terms to balance the distribution of such terms in toxic comments and non-toxic comments.

[3] carry out several investigations into identity term bias in a BERT-based classifier for TCC tasks. They find that BERT is over-attentive to identity terms and neglecting the context around the identity terms, which contributes to the false positive errors. To handle the bias, [3] propose BERT+SOC (Sampling and Occlusion), which is built atop BERT with an extra regularization term to regularize the importance scores of identity terms assigned by SOC. The basic idea

behind this is to minimise the prediction differences between when an identity term is exposed to the model and when it is hidden from the model. Ideally, the over-attended identity terms will be assigned with low importance scores by SOC and thus they will become less indicative of whether the comment is hate speech or not.

### 2.3 Remarks

In summary, despite a significant amount of works on TCC, existing methods have shown to make biased predictions based on identity terms. This has become a focus of studies in recent years. Existing methods to address this follow a similar principle that encourages the model to ignore or pay less attention to identity terms. However, this overlooks the important fact that in particular situations, such terms are useful for the prediction.

In this work, we explore a new venue for addressing the identity term bias in TCC models. We design the model to consider the subjectivity level of the whole comment when an identity term appears in the comment. The phrase *subjectivity level* is used here to describe to what extent the comment conveys personal opinion or factual information. A comment of a high level of subjectivity indicates the comment contains more personal opinion and less factual information, vice versa. We demonstrate this with an in-depth analysis below.

## 3 Analyzing bias towards identity terms

In most types of toxic comments, e.g. hate speech, aggressive language and abusive language, they are likely to express hate or encourage violence towards a person or group based on certain characteristics such as race, religion, sex, or sexual orientation [20]. Such expressions are intuitively more a reflection of personal feelings rather than fact-quoting. Therefore in the following, we analyse some commonly used toxic comment datasets to quantify this.

### 3.1 The task and datasets

We include four datasets in the analysis, with the aim to cover different social media platforms, dataset sizes and text lengths. Since the identity term bias is found in various types of toxic content and also to follow the practice by [3] which study the identity term bias in the context of binary classification, we group different toxic comments into one group without distinguishing their specific types. Therefore, the task in this work is a binary toxic comment classification task where the model aims to predict if the comment is toxic or not.

The first dataset is collected from a white supremacist online forum (*WS*) [21]. We select this dataset as it is employed by [3] to study identity term bias, thus using the same dataset can allow fair comparisons. It includes 10,703 posts in total and 1,196 of them are "Toxic", 9,507 are "non-Toxic" [1]. The second and third datasets are both collected from Twitter. For the second dataset, denoted as *Twitter 18k* [8], we convert the labels "Racism", "Sexism" and "Both" to "Toxic" and consider the label "Neither" as "non-Toxic". The third dataset is denoted as *Twitter 42k* [22]. In the same vein, labels of "Abusive" and "Hateful" are converted into "Toxic", and "Normal" is treated as "non-Toxic". We remove the examples of label "Spam" as they are not a type of toxic comments according to the typology of "hate-based rhetoric" proposed in [23]. After conversion, Twitter 18k contains 18,625 tweets in total with 5,814 "Toxic" tweets and 12,811 "non-Toxic" tweets. Twitter 42k has 5,705 tweets as toxic and 36,609 as normal[2]. The fourth dataset is collected from Wikipedia Talk page and annotated in a multi-label classification approach, denoted as *Wiki* [24]. There are six labels in total, namely "toxic", "severe toxic", "threat", "obscene", "insult" and "identity hate". We convert the label of a post to "Toxic" if the post has at least one of the six labels, and convert the remaining posts to "non-Toxic". This leads to 16,225 posts with "Toxic" labels and 143,346 with "non-Toxic" labels.

In short, the four selected datasets contain between 15,000 and 159,571 comments and cover three different social media platforms with different average comment lengths. Table 1 as shown below summarises the four datasets.

### 3.2 Defining "Subjectivity"

A few studies on TCC have utilized the subjectivity-related features or directly focused on detecting subjectivity contents [27, 20, 28, 29, 30]. However, there is a lack of clear definition of subjectivity or consensus on how this should be formally defined. For example, [20] points out that "a subjective sentence expresses some feelings, views, or beliefs." Other studies, such as [31] and [32], directly focus on subjectivity detection but none of them gives an actual definition of subjectivity or an explanation of what kind of comments or texts are supposed to be labeled as "subjective". We adopts the definition of "subjectivity level" by TextBlob [33] as mentioned before: that subjectivity level describes the extent to which the comment conveys personal opinion or factual information. A comment with a high level of subjectivity indicates the comment contains more personal opinion and less factual information, vice versa.

---

[1] The original binary labels are "hate" and "no hate".
[2] The original version dataset includes over 80,000 tweets with their tweets IDs and labels published [22]. We have retrieved 50,425 valid tweets using their tweets IDs. The remaining tweets have been deleted as by the point of this study. After removing "spam" tweets, 42,314 tweets are kept for this dataset.

| Dataset | Source | Data Numbers | Original Labels | Toxic Proportion | Avg Text Length |
|---|---|---|---|---|---|
| WS [25] | Stormfront | 14,998 | non-aggressive (42%), overtly aggressive (35%) covertly aggressive (23%) | 11.17% | 91 |
| Tweet 18k[26] | Twitter | 18,625 | racism (11%), sexism (20%), both (69%), neither | 31.22% | 96 |
| Tweet 42k[22] | Twitter | 50,425 | abusive (8%), hateful (3%), normal (73%), spam (16%) | 13.48% | 123 |
| Wiki [24] | Wikipedia | 159,571 | toxic (10%), severe toxic (1%), obscene (5%), threat (0.3%), insult (5%), identity hate (1%) | 10.17% | 398 |

Table 1: Summary of the four toxic comment classification tasks. "Toxic Proportion" refers to the proportion of "Toxic" comments after the conversion to binary classification.

The TextBlob[3] library [33] is the tool this work uses to generate subjectivity scores to facilitate our bias analysis. We have also identified the other similar tool, SentiWordNet[34]. Given a text, both tools assign a subjectivity score within the range from 0.0 to 1.0 where 0.0 is very objective and 1.0 is very subjective. However, we conducted an analysis that included manually inspecting and validating the output of the two tools on a sample of 80 data instances from the 4 datasets. We believe that TextBlob is more accurate at predicting subjectivity. This is because, first, as shown in Figure 1, SentiWordNet assigns most text an extreme score of '1' (very subjective), while TextBlob is more sensitive to the subjectivity levels over the four datasets. We also found these two patterns during our manual investigation.

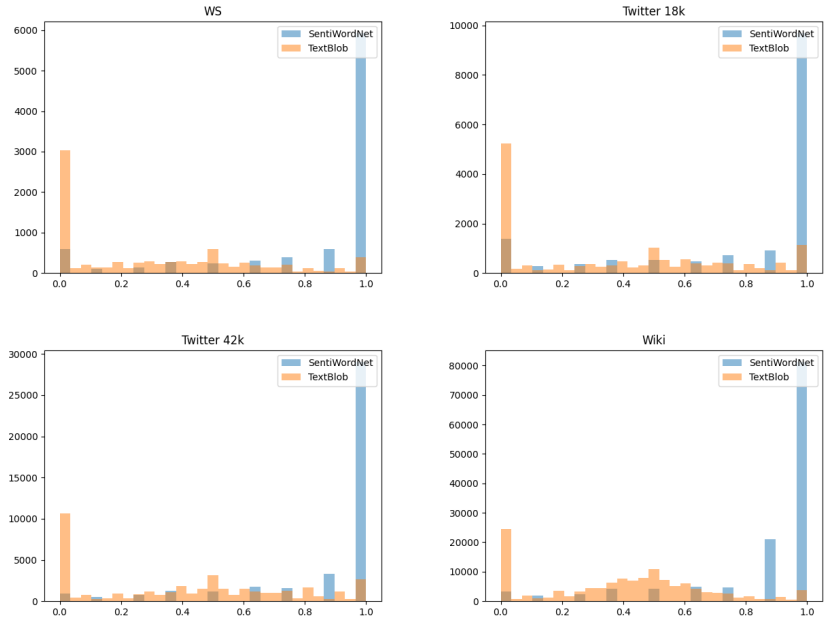

Figure 1: The subjective scores distribution of TextBlob and SentiWordNet over the four datasets. The X-axis is the comment count of corresponding subjectivity scores indicated by the Y-axis. All four plots are based on the training dataset.

Second, we conduct a qualitative analysis over the comments that these two tools have a disagreement greater than 0.5. To given an example, the following comment:

*"my mothers father is full blonde Irish. my mothers mother is English Irish Scottish . my fathers father is a German with some Scottish and my fathers mother is danish and Norwegian. making me 6ft4 blonde hair blue eyes with a big barrel chest just like my german grandfather ."*

is scored differently by TextBlob(*0.0682*) and SentiWordNet(*1*). In this case, We take the score *0.0682* by TextBlob as the more accurate one. We randomly take out 20 comments with such subjectivity scores disagreements from each dataset. We end up with 80 comments and we our manual investigation shows that TextBlob is more accurate for 77.5% sample we validated.

---
[3] https://textblob.readthedocs.io/en/dev/#

Therefore, we believe TextBlob is more accurate and more sensitive on these four datasets. The more refined granularity probably is a more accurate reflection of subjectivity in the data. It will also offer the model more information on the subjectivity level of comments.

### 3.3 Predictive model: BERT

Influenced by [3], we use BERT as the classifier to demonstrate the problem of identity term bias in TCC tasks. Another reason for choosing BERT as the model to analyse the bias is that it is widely used in various NLP tasks and many pre-trained language models are built upon it [15, 17, 16].

In short, given a fixed length token sequence[4], BERT assigns each token an initial embedding and an attention mask. The pre-trained embeddings are supposed to contain the semantics information for the token. The attention mask of a binary value (*0* or *1*) indicates if the token is a padded token or not to avoid the model performing attention on padded tokens. To be more specific, between the attention mask values, 0 is for tokens that are padded for making up to the fixed-length input so they are masked and the model does not need to attend to them during the training, 1 is for the actual tokens from input sentences which the model attends to. Embeddings and masks are fed into the Transformer Encoder layers and the final logit is used for the loss calculation and for the final classification. We train the initial BERT model on the four TCC tasks separately and investigate their prediction results. We follow the implementation details used in [3], which are further explained in Section 5.

### 3.4 Bias in model predictions: qualitative analysis

| Identity Terms | Predictions | Comments | Subjectivity scores |
|---|---|---|---|
| muslim | True Positive (toxic) | i have been talking about it for years. the filthy **muslims** putting their sight on africa. in some ways they are worst then blacks. | 1 |
| | False Positive (non-toxic) | @ kelly turn er 9 9 @ bu eller ish ere more **muslims** die in terrorist attacks than anyone else. for example, 80% of afghan civilians killed by taliban | 0 |
| gay | True Positive (toxic) | if this guy isn t gay he is one seriously good actor !!!!! i've seen hayes on will & grace and the way he acts so.... feminine he must be homosexual. no straight man can act so convincing ly **gay**. but then again sean acted pretty straight and normal in the movie where he played jerry lewis. my opinion is still in favour of him being gay though. | 0.6444 |
| | False Positive (non-toxic) | f yi i am a **gay** man , if anyone wants to chat just comment pl z | 0.3417 |
| whites | True Positive (toxic) | black people in sa do not want to unite with **whites** to remove corruption . we ' d rather endure rot & amp; suffer! so sad! | 0.6444 |
| | False Positive (non-toxic) | i' m reporting here that i haven' t seen or heard anything on talk radio or tv that even hints at jewish su pre mac ist ethnic cleansing of **whites** from usa or europe . | 0 |
| women | True Positive (toxic) | rt @ mwilbur62 i swear i am not sex ist , but **women** ' s tennis is the most boring sport to watch | 0.75 |
| | False Positive (non-toxic) | men and **women** are segregated in mosques . | 0 |
| jews | True Positive (toxic) | well i am sure that the **jews** have the cure for aids but they won t give it to the fa gs cause they already served their pu rp oise | 0.8888 |
| | False Positive (non-toxic) | if a third world savage attacks a **jew**, is it all jew ey ' s fault? | 0 |

Table 2: Comment examples that contain identity terms. Each pair of comments contain the same identity terms, while one is true positive and the other is false positive predicted by BERT.

Errors made by BERT are extracted for our analysis, i.e. false positive predictions and false negative predictions. As mentioned before, we use the TextBlob to generate a subjectivity score for those comments [33].

We investigate those errors and select several representative samples with identity terms as shown in Table 2. We observe that to correctly predict the toxic comments with identity terms, we need to account for the meaning of the whole sentence and the stance of the speakers. Subjectivity level is possibly one perspective from which this can be captured. For example, in the first pair which mentions "muslims" in Table 2, the toxic comment compares "blacks" with "muslims" and asserts that "they (Muslims) are worst than blacks" without any factual reasoning, which indicates a high level of subjectivity. While the non-toxic comment with the term "muslims" attempts to convey an objective fact related to Muslims, although it contains comparison, it attempts to include a specific figure to support the comparison[5].

---

[4]One word is converted to one or multiple tokens.

[5]The "fact" and figure the comment provides are unnecessarily true but it is not the topic we aim to study.

This observation inspires us that the subjectivity score of the comments can be a helpful indicator when classifying a comment with identity terms.

### 3.5 Bias in model predictions: quantitative analysis

To further analyse the subjectivity level at scale, we conduct a quantitative analysis of the subjectivity score of false positive and true positive predictions. We separate those comments with identity terms and those without identity terms to examine the identity term bias. The identity terms list is adopted from [3] which includes 25 terms[6] such as "muslim", "jew", and "women".

In this way, all comments are essentially split into four groups: true positive with identity terms (TPwIT), false positive with identity terms (FPwIT), true negative without identity terms (TNwoIT), false negative without identity terms (FNwoIT). We plot the subjectivity score distribution over false positives and true positives with box-plot diagrams. As shown on the left of Figure 2, for comments with identity terms, the true positives show higher subjectivity levels than the false positives across all tasks. First, false positives have a lower median of subjectivity scores than true positives across four datasets. Second, the false positive predictions have a generally smaller and lower interquartile range than the true positive predictions in the task. The lower subjectivity scores in false positive predictions may reflect the real-word scenario that when speakers talk about a demographic group such as female, Muslim or Asian in an objective way, e.g. describing the group neutrally, the speech is less likely to be disrespectful or offensive. On the other hand, toxic comments often involve subjective expressions.

Notably, the pattern of lower subjectivity level of false positives is consistent only among comments with identity terms. The comments without identity terms, as shown on the right diagram in Figure 2 do not indicate a consistent pattern between false positives and true positives.

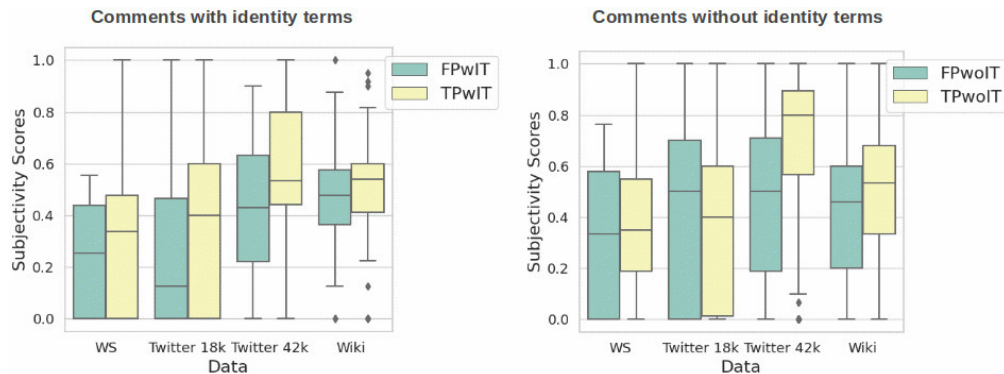

Figure 2: The comparison of subjectivity scores between true positive (i.e. toxic comments, as coloured in yellow) and false positive predictions (i.e. non-toxic comments, as coloured in green) by an initial BERT classifier. This figure is better viewed in colour.

### 3.6 Remark

In summary, using the predictions by a state of the art BERT classifier, we observe that when identity terms are present, false positives tend to have lower subjectivity scores while true positives generally have higher subjectivity scores. Based on this observation, we hypothesize that the subjectivity score of a comment along with the presence of identity terms can be useful in classifying toxicity. We introduce our method to leverage this in the next section.

## 4 Subdentity-BERT (SS-BERT)

As analysed in Section 3, among comments with identity terms, false positive predictions from BERT, i.e. non-toxic comments, tend to have lower subjectivity scores. Given this, we design our model such that it pays attention to the subjectivity of a comment when the comment contains identity terms. When the identity terms are not present, the model should not consider the subjectivity.

---

[6]The full list are: muslim,,jew, jews, white, islam, blacks, muslims, women, whites, gay, black, democat, islamic, allah, jewish, lesbian, transgender, race, brown, woman, mexican, religion, homosexual, homosexuality, africans.

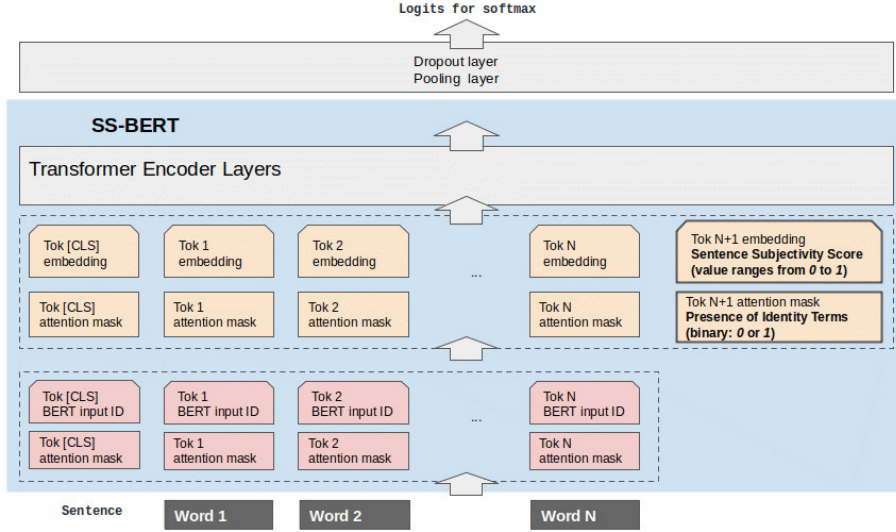

Figure 3: Illustration of a SS-BERT model on classification tasks. This figure is better viewed in colour.

To do so, building on BERT, we append an additional "token" to the end of the token sequence for each comment as shown in Figure 3. We use the subjectivity score for the embedding of the token. To be more specific, we create a 3-D tensor with the same size of other token embeddings and each element's value in the tensor is equal to the subjectivity score. In the scenario of BERT, the dimension size is 768 and thus the tensor for the added "token" is a 3D-tensor of size [batch size, 1, 768]. The tensor is denoted as "Sentence Subjectivity Score" in Figure 3. For the corresponding attention mask (highlighted in yellow with bold borders in Figure 3), we set it to indicate the presence of identity terms so that if there is no identity term in the comment, the appended "token" will be masked. While if there is an identity term, the embedding of this "token", i.e. the subjectivity score, will be attended by BERT.

## 5 Experiment

### 5.1 Comparative models

**Baselines** In this work, we set up two baselines. The first baseline is an initial BERT, as explained in Section 3.3. The second baseline, BERT+SOC, is an implementation of the method in [3], as described in Section 2.2 which is a SOTA model also focusing on the identity term bias.

**Subjectivity-Only BERT (SO-BERT)** To examine if the subjectivity level is especially helpful for the identity term bias, we conduct an ablation study by testing a model, SO-BERT, which only captures the information of subjectivity level but not the presence of identity terms. In other words, the SO-BERT model "considers" the subjectivity level for a comment regardless of whether it contains identity terms or not. To do so, we adapt from SS-BERT model with the attention mask always attending the added "token". The rest of the model structure remains the same.

**Subjectivity-BERT+SOC (SS-BERT+SOC)** Additionally, we combine the method in [3] (BERT+SOC, as introduced in Section 2) and our model (SS-BERT) to create another BERT-based model, SS-BERT+SOC. In short, this hybrid model learns the subjectivity level and the presence of identity terms with the added "token" and also has an extra regularization term in its loss function which encourages the model to learn more from the context of the identity term and less from the identity term.

In short, we compare SS-BERT with two baseline models and two models which also incorporate subjectivity levels but with different methods from SS-BERT. We empirically create the other two models for comparisons. Our assumption is that SS-BERT, by considering both the subjectivity level and identity terms, will perform the best.

### 5.2 Datasets

We evaluate each model on 4 different datasets as introduced in Section 3.1. Each dataset is split to training, validation and test datasets (80%, 10%, 10%). The results reported in this work are all on the testing dataset.

### 5.3 Implementation

We constrain the maximum length of each input instance to be 128 tokens for WS, Twitter 18k and Twitter 42k, and 400 tokens for Wiki. The maximum length is set up based on the average length of text in each dataset as detailed in Table 1 in the appendix.

The hyperparameter settings follow those in [3]. Accordingly, the batch size is set to 32. Adam optimization is implemented with a starting learning rate of 2 x 10-5. The validation is performed every 200 steps and the learning rate is halved every time the validation F1 decreases. The model stops training after the learning rate halved 5 times. We also re-weight the training loss to handle the imbalance labels as [3].

We used a single NVIDIA Tesla V100 GPU for all experiments. We run the two baseline models (BERT and BERT+SOC) with the code provided by [3]. Paying tributes to [3], our code for SS-BERT, SS-BERT+SOC and SO-BERT is partly adapted from theirs, such that our implementation uses the same software packages as their work [7]. For each task, a model runs 10 times independently to give a mean F1 score, following the implementation in [3].

### 5.4 Results: F1 comparison

| DATA | Total data size | BERT | | BERT+SOC | | SS-BERT | | SS-BERT+SOC | | SO-BERT | |
|---|---|---|---|---|---|---|---|---|---|---|---|
| | | F1 | std | F1 | std | F1 | std | F1 | std | F1 | std |
| WS | 10,703 | 0.5811 | 0.0204 | 0.5885 | 0.0209 | **0.5952** | 0.0203 | 0.5912 | 0.0216 | 0.5909 | 0.0247 |
| Twitter 18k | 18,625 | 0.7780 | 0.0204 | 0.7780 | 0.0055 | **0.7804** | 0.0080 | 0.7785 | 0.0050 | 0.7774 | 0.0055 |
| Twitter 42k | 42,314 | 0.7637 | 0.0071 | 0.7643 | 0.0101 | **0.7683** | 0.0059 | 0.7660 | 0.0056 | 0.7636 | 0.0061 |
| Wiki | 159,571 | 0.7680 | 0.0175 | 0.7548 | 0.0135 | **0.7693** | 0.0133 | 0.7568 | 0.0112 | 0.7654 | 0.0151 |

Table 3: A comparison of F1 scores on 4 TCC tasks. The mean F1 score and its standard deviation are from 10 independently runs for each model presented.

Table 3 shows the F1 score of each model on 4 datasets. Overall, model SS-BERT achieves the best performances consistently across 4 datasets and it is also the only model that outperforms the baseline BERT consistently across 4 datasets.

Between the two baselines, BERT and BERT+SOC[8], except on the biggest dataset Wiki, BERT+SOC is able to improve BERT's performance on TCC tasks. This partly reflects the results reported in [3]. The under-performance of BERT+SOC on Wiki may indicate that BERT could have benefited from training on the significantly larger dataset (compared to WS, Twitter 18k and Twitter 42k) such that the extra learning objective for learning the context had negligible influence on the model. In contrast, our model SS-BERT outperforms BERT and BERT+SOC on all datasets. This is because the mechanism of attending to subjectivity based on the presence of identity terms cannot be compensated by dataset size. Therefore, the results show that our model brings unique benefits to the two SOTA models and that is the reason SS-BERT still outperforms BERT on Wiki.

Among the three models of BERT+SOC, SS-BERT and SS-BERT+SOC, SS-BERT consistently achieves the best performance on all datasets, with SS-BERT+SOC achieving the second and BERT+SOC the third on all datasets. This indicates that BERT+SOC and SS-BERT handle the identity term bias using different mechanisms. Although BERT+SOC is designed to mitigate the identity term bias, it is not able to learn the subjectivity level of comments with identity terms. Therefore, SS-BERT+SOC outperforms BERT+SOC consistently, suggesting that adding the subjectivity level and the presence of identity terms can improve BERT+SOC performance. However, SS-BERT+SOC does not exceed SS-BERT. A possible explanation can be that the extra regulation from SOC might dilute the impact that SS-BERT brings to the model.

Notably, SO-BERT, the model that only considers subjectivity regardless of the presence of identity terms, does not benefit TCC tasks. It only achieves better performance than BERT on WS. This reflects the intuition we mentioned previously and the pattern we identified in Section 3 that subjectivity are indicative features of toxicity only if identity terms are present in a comment.

| DATA | BERT | | SS-BERT | |
|---|---|---|---|---|
| | False Positive | False Negative | False Positive | False Negative |
| WS | 34 | 57 | 34 | 55 |
| Twitter 18k | 149 | 117 | 146 | 117 |
| Twitter 42k | 227 | 78 | 214 | 81 |
| Wiki | 750 | 168 | 738 | 170 |

Table 4: Summary of false positives and false negatives of BERT and SS-BERT on the 4 datasets. Mean values of the performance across 10 runs are shown.

### 5.5 Results: identity term bias

We compare the erroneous predictions between SS-BERT and the baseline BERT to further investigate SS-BERT's performance, especially regarding tackling false positive bias and identity term bias. First, as shown in Table 4, SS-BERT is able to decrease the false positive predictions across all tasks. This reflects our hypothesis that it is helpful to mitigate the identity term bias, i.e. false positive bias, by letting the model learn the subjectivity level of a comment when an identity term appears. The dataset Twitter 42k benefits from this the most, given its dataset size and the decreased false positives. This is consistent with Figure 2 which shows a noticeable gap of subjectivity levels between false positive and true positive predictions from BERT on Twitter 42k.

### 5.6 Error Analysis

We perform an error analysis to shed light on the limitations of the best performing model (SS-BERT) on TCC with a focus on the identity term bias.

First, the identity term list we use may not "fit" the datasets well. An updated identity term list may bring further improvement. Specifically, there is only a small percentage of comments that contain the identity terms used in the model. For WS, Twitter 18k, Twitter 42k and Wiki, the ratio of comments with identity terms in the list are 21.20%, 19.90%, 4.22%, and 6.02% respectively. Therefore, only a small percentage of the datasets could have benefited from our method. For example, we observe that some false negatives and false positives contain identity terms like "liberal", "irish", "asian", and "chinese", which are not included in the identity term list used in our model. Examples are listed in Table 5 as shown below. [3], they use a supervised approach combined with manual screening to select identity terms. One approach to address the issue is creating an universal list of identity terms that covers as many identity terms as possible.

| Comments | Group identity terms | SS-BERT Predictions |
|---|---|---|
| liberal is just the pc word for rap ist . | liberal | False Negative |
| i get what your saying and there are thousands of non irish here who will never contribute anything to this land . | irish | False Negative |
| i suppose the concept of a male cow and a female cow producing more cows , and the repeating process is too much for them to understand . | female | False Negative |
| traditionally , there have always been a lot of them despite the fact that ne asian s are the extreme version of our breeding strategy . | asian | False Negative |
| there is no doubt the chinese and ja ps have giving the world culture . | chinese | False Positive |
| the reason they are killing themselves is because our weak liberal gov er ment are too afraid to make them do any sort of training . . | liberal | False Positive |
| she came to hate the african po ache rs and showed no mercy towards them . | african | False Positive |
| these girls are not there for the cooking  mk r what a joke . | girls | False Positive |

Table 5: Examples of false positives and false negatives with group identifier terms that not included in the identity term list used in our model.

Second, the subjectivity scores given by TextBlob are not always accurate for each comment with identity terms. We believe accurate subjectivity scores can potentially help SS-BERT predict them more accurately. For example, the false positive *"We will see what happens . I just hope this does not cause Muslims hate . Extremist if they did it"* has been assigned a high subjectivity score of 0.9. More examples of this kind are shown in Table 6 as below.

---

[7] Therefore, we use "bert-base-uncased" BERT as [3].

[8] We notice that our results of BERT and BERT+SOC on the WS dataset are different from that reported in [3], as our F1 are higher. While our results are obtained by re-running their code as-is, a possible reason for this difference is that the only version of the data we can download has been modified from that used in the authors' original study.

| Comments | Subjectivity Scores | SS-BERT Prediction |
|---|---|---|
| @ mad dr 11 there is not and never has been a country called palestine , micro bra in . and that land has been the home of jews for 3000 years . | 1 | False Positive |
| buying black jelly beans for my kids . . . in case they re cutting up and need a chill pill . they hate them . | 0.7444 | False Positive |
| i ll never understand how guys cheat on gorgeous women that they shouldn t have gotten a chance with to begin with with trolls . | 0.9000 | False Positive |
| we will see what happens . i just hope this does not cause muslims hate . ex tre mist if they did it | 0.9000 | False Positive |
| as far as muslims are concerned any criticism of islam is hate speech . i have been a muslim for five years and know perfectly the hatred that muslims have for ap osta tes and the level of into ler ance towards their critics . been there , done that myself , without even being aware of it . it all seemed normal to do . today i find myself on the other side of the fence . quoting opinions contrary to those believed by muslims is not hate speech . the quran is filled with hate speech . look at the article ali sin a for example : that is hate speech . the very language is despite ful . my every contribution to islam | 0.7341 | False Positive |
| why white people used to say that sex was a sin used to be a mystery to me until i saw the children of browns and mixed race children popping up all around me . | 0.1250 | False Negative |
| saudi cleric believes earth does not rev ol ve around sun . there must be 50 ways to return to the 7th century with islam as your expert guide | 0 | False Negative |
| rt @ ashes cho w : women ' s studies : learn how to tell other women they should study something other than women ' s studies rt @ eli ah $_r$ : $http://t...$ | 0.3750 | False Negative |
| no wonder this races achieve nothing . | 0 | False Negative |
| lack of compassion seems to go hand and hand with lack the of white blood . | 0 | False Negative |

Table 6: Examples of false positives assigned with high subjectivity scores and false negatives assigned with low subjectivity scores.

### 5.7 Generalization

To examine the generalization of our hypothesis, we also test our method on RoBERTa, which is another often-used pre-trained LM for TCC tasks [17, 35]. RoBERTa has a similar structure with BERT where each token is given an embedding and an attention mask [17]. We add the information of subjectivity scores to the embeddings and the presences of identity terms to attention masks, as exactly how we do with SS-BERT. To provide a fair comparison, we use an initial RoBERTa as the baseline. The implementations are similar to how we implement BERT-based models that details are shown in the appendix.

The RoBERTa model and the SS-RoBERTa model are built on our code of BERT and SS-BERT. We use a RoBERTa of "roberta-base". The hyperparameter settings dataset processing are the same as the experiments of BERT-based models.

The results (Table 7) show that our method consistently improves the baseline RoBERTa across 4 datasets with the maximum improvement up to 1.29%.

Table 7: Performance comparison between SS-RoBERTa and RoBERTa. The mean F1 score and its standard deviation, the mean false negative and the mean false positive are from 10 independently runs for each model presented. FN: False Negative; FP: False Positive

## 6 Conclusion

In this paper, we proposed a novel approach to tackle the identity term bias problem in TCC tasks. Our approach is mainly based on paying additional attention to the subjectivity level of comments when an identity term appears. Our model SS-BERT outperforms SOTA methods on a wide range of TCC tasks. The results reveal that our method can mitigate the bias toward identity terms and reduce the false positive predictions effectively.

Our future work will look to address the limitations discussed before, i.e., developing an extensive identity term list and addressing inaccuracies in computing subjectivity. Another issue not addressed in this study and can be a possible direction is generalising our method to other pre-trained models which have different structures from BERT, such as Transformer-XL that does not include attention masks dai2019transformer.

## References

[1] Ji Ho Park, Jamin Shin, and Pascale Fung. Reducing gender bias in abusive language detection. In *Proceedings of the 2018 Conference on Empirical Methods in Natural Language Processing*, pages 2799–2804, 2018.

[2] Lucas Dixon, John Li, Jeffrey Sorensen, Nithum Thain, and Lucy Vasserman. Measuring and mitigating unintended bias in text classification. In *Proceedings of the 2018 AAAI/ACM Conference on AI, Ethics, and Society*, pages 67–73, 2018.

[3] Brendan Kennedy, Xisen Jin, Aida Mostafazadeh Davani, Morteza Dehghani, and Xiang Ren. Contextualizing hate speech classifiers with post-hoc explanation. In *Proceedings of the 58th Annual Meeting of the Association for Computational Linguistics*, pages 5435–5442, 2020.

[4] Irene Kwok and Yuzhou Wang. Locate the hate: Detecting tweets against blacks. In *Twenty-seventh AAAI conference on artificial intelligence*, 2013.

[5] Vikas S Chavan and SS Shylaja. Machine learning approach for detection of cyber-aggressive comments by peers on social media network. In *2015 International Conference on Advances in Computing, Communications and Informatics (ICACCI)*, pages 2354–2358. IEEE, 2015.

[6] Pete Burnap, Omer F Rana, Nick Avis, Matthew Williams, William Housley, Adam Edwards, Jeffrey Morgan, and Luke Sloan. Detecting tension in online communities with computational twitter analysis. *Technological Forecasting and Social Change*, 95:96–108, 2015.

[7] Paula Fortuna and Sérgio Nunes. A survey on automatic detection of hate speech in text. *ACM Computing Surveys (CSUR)*, 51(4):85, 2018.

[8] Zeerak Waseem and Dirk Hovy. Hateful symbols or hateful people? predictive features for hate speech detection on twitter. In *Proceedings of the NAACL student research workshop*, pages 88–93, 2016.

[9] Ji Ho Park and Pascale Fung. One-step and two-step classification for abusive language detection on twitter. In *Proceedings of the First Workshop on Abusive Language Online*, pages 41–45, 2017.

[10] Betty van Aken, Julian Risch, Ralf Krestel, and Alexander Löser. Challenges for toxic comment classification: An in-depth error analysis. In *Proceedings of the 2nd Workshop on Abusive Language Online (ALW2)*, pages 33–42, 2018.

[11] Anna Schmidt and Michael Wiegand. A survey on hate speech detection using natural language processing. In *Proceedings of the Fifth International Workshop on Natural Language Processing for Social Media*, pages 1–10, 2017.

[12] Fabio Del Vigna, Andrea Cimino, Felice Dell'Orletta, Marinella Petrocchi, and Maurizio Tesconi. Hate me, hate me not: Hate speech detection on facebook. In *Proceedings of the First Italian Conference on Cybersecurity (ITASEC17)*, pages 86–95, 2017.

[13] Zichao Yang, Diyi Yang, Chris Dyer, Xiaodong He, Alex Smola, and Eduard Hovy. Hierarchical attention networks for document classification. In *Proceedings of the 2016 Conference of the North American Chapter of the Association for Computational Linguistics: Human Language Technologies*, pages 1480–1489, 2016.

[14] Ziqi Zhang, David Robinson, and Jonathan Tepper. Detecting hate speech on twitter using a convolution-gru based deep neural network. In *European semantic web conference*, pages 745–760. Springer, 2018.

[15] Jacob Devlin, Ming-Wei Chang, Kenton Lee, and Kristina Toutanova. Bert: Pre-training of deep bidirectional transformers for language understanding. In *Proceedings of the 2019 Conference of the North American Chapter of the Association for Computational Linguistics: Human Language Technologies, Volume 1 (Long and Short Papers)*, pages 4171–4186, 2019.

[16] Guillaume Lample and Alexis Conneau. Cross-lingual language model pretraining. *arXiv preprint arXiv:1901.07291*, 2019.

[17] Yinhan Liu, Myle Ott, Naman Goyal, Jingfei Du, Mandar Joshi, Danqi Chen, Omer Levy, Mike Lewis, Luke Zettlemoyer, and Veselin Stoyanov. Roberta: A robustly optimized bert pretraining approach. 2019.

[18] Marzieh Mozafari, Reza Farahbakhsh, and Noël Crespi. A bert-based transfer learning approach for hate speech detection in online social media. In *International Conference on Complex Networks and Their Applications*, pages 928–940. Springer, 2019.

[19] Zhixue Zhao, Ziqi Zhang, and Frank Hopfgartner. A comparative study of using pre-trained language models fortoxic comment classification. In *Companion Proceedings of the Web Conference 2021*, 2021.

[20] Njagi Dennis Gitari, Zhang Zuping, Hanyurwimfura Damien, and Jun Long. A lexicon-based approach for hate speech detection. *International Journal of Multimedia and Ubiquitous Engineering*, 10(4):215–230, 2015.

[21] Ona de Gibert, Naiara Perez, Aitor García-Pablos, and Montse Cuadros. Hate speech dataset from a white supremacy forum. In *Proceedings of the 2nd Workshop on Abusive Language Online (ALW2)*, pages 11–20, 2018.

[22] Antigoni Maria Founta, Constantinos Djouvas, Despoina Chatzakou, Ilias Leontiadis, Jeremy Blackburn, Gianluca Stringhini, Athena Vakali, Michael Sirivianos, and Nicolas Kourtellis. Large scale crowdsourcing and characterization of twitter abusive behavior. In *Twelfth International AAAI Conference on Web and Social Media*, 2018.

[23] Brendan Kennedy, Drew Kogon, Kris Coombs, Joe Hoover, Christina Park, Gwenyth Portillo-Wightman, Aida Mostafazadeh, Mohammad Atari, and Morteza Dehghani. A typology and coding manual for the study of hate-based rhetoric, 2018.

[24] ConversationAI. Toxic comment classification challenge: Identify and classify toxic online comments, 2017.

[25] Ritesh Kumar, Aishwarya N. Reganti, Akshit Bhatia, and Tushar Maheshwari. Aggression-annotated Corpus of Hindi-English Code-mixed Data. In Nicoletta Calzolari (Conference chair), Khalid Choukri, Christopher Cieri, Thierry Declerck, Sara Goggi, Koiti Hasida, Hitoshi Isahara, Bente Maegaard, Joseph Mariani, Hélène Mazo, Asuncion Moreno, Jan Odijk, Stelios Piperidis, and Takenobu Tokunaga, editors, *Proceedings of the Eleventh International Conference on Language Resources and Evaluation (LREC 2018)*, Miyazaki, Japan, May 7-12, 2018 2018. European Language Resources Association (ELRA).

[26] Zeerak Waseem. Are you a racist or am i seeing things? annotator influence on hate speech detection on twitter. In *Proceedings of the First Workshop on NLP and Computational Social Science*, pages 138–142, Austin, Texas, November 2016. Association for Computational Linguistics.

[27] Matthew Purver and Stuart Battersby. Experimenting with distant supervision for emotion classification. In *Proceedings of the 13th Conference of the European Chapter of the Association for Computational Linguistics*, pages 482–491. Association for Computational Linguistics, 2012.

[28] Cynthia Van Hee, Gilles Jacobs, Chris Emmery, Bart Desmet, Els Lefever, Ben Verhoeven, Guy De Pauw, Walter Daelemans, and Véronique Hoste. Automatic detection of cyberbullying in social media text. *PloS one*, 13(10):e0203794, 2018.

[29] Reid Pryzant, Richard Diehl Martinez, Nathan Dass, Sadao Kurohashi, Dan Jurafsky, and Diyi Yang. Automatically neutralizing subjective bias in text. In *Proceedings of the aaai conference on artificial intelligence*, volume 34, pages 480–489, 2020.

[30] Jan Kocoń, Alicja Figas, Marcin Gruza, Daria Puchalska, Tomasz Kajdanowicz, and Przemysław Kazienko. Offensive, aggressive, and hate speech analysis: From data-centric to human-centered approach. *Information Processing & Management*, 58(5):102643, 2021.

[31] Chenghua Lin, Yulan He, and Richard Everson. Sentence subjectivity detection with weakly-supervised learning. In *Proceedings of 5th International Joint Conference on Natural Language Processing*, pages 1153–1161, 2011.

[32] Hairong Huo and Mizuho Iwaihara. Utilizing bert pretrained models with various fine-tune methods for subjectivity detection. In *Asia-Pacific Web (APWeb) and Web-Age Information Management (WAIM) Joint International Conference on Web and Big Data*, pages 270–284. Springer, 2020.

[33] Steven Loria. textblob documentation. *Release 0.15*, 2, 2018.

[34] Stefano Baccianella, Andrea Esuli, and Fabrizio Sebastiani. Sentiwordnet 3.0: an enhanced lexical resource for sentiment analysis and opinion mining.

[35] Arup Baruah, Kaushik Das, Ferdous Barbhuiya, and Kuntal Dey. Aggression identification in english, hindi and bangla text using bert, roberta and svm. In *Proceedings of the Second Workshop on Trolling, Aggression and Cyberbullying*, pages 76–82, 2020.